\documentclass[a4paper,10pt]{article}
\usepackage[utf8]{inputenc}
\usepackage{amsmath, amssymb, amsthm}
\usepackage{graphicx,subcaption}
\usepackage[sorting=none]{biblatex}
\usepackage{tabularx}
\usepackage{wrapfig}
\bibliography{references.bib}
\usepackage{multirow}
\usepackage{float}
\newcommand{\cmmnt}[1]{}
\usepackage{filecontents}

\title{Out-of-distribution detection algorithms for robust insect classification}

\author
{Mojdeh Saadati$^{1}$, Aditya Balu$^{2}$, Shivani Chiranjeevi$^{2}$, Talukder Zaki Jubery$^{2}$,\\ Asheesh K Singh$^{3}$, Soumik Sarkar$^{1,2}$,  Arti Singh$^{3\ast},$ Baskar Ganapathysubramanian$^{2\ast}$ \\
\\
\normalsize{$^{1}$Department of Computer Science, Iowa State University, Ames, IA, USA}\\
\normalsize{$^{2}$Department of Mechanical Engineering, Iowa State University, Ames, IA, USA}\\
\normalsize{$^{3}$Department of Agronomy, Iowa State University, Ames, IA, USA}\\
\normalsize{$^\ast$To whom correspondence should be addressed; E-mail:  baskarg@iastate.edu, arti@iastate.edu}
\\
}

\begin{document}
\maketitle
\begin{abstract}
Plants encounter a variety of beneficial and harmful insects during their growth cycle. Accurate identification (i.e., detecting insects' presence) and classification (i.e., determining the type or class) of these insect species is critical for implementing prompt and suitable mitigation strategies. Such timely actions carry substantial economic and environmental implications. Deep learning-based approaches have produced models with good insect classification accuracy; Most of these models are conducive for application in controlled environmental conditions. One of the primary emphasis of researchers is to implement identification and classification models in the real agriculture fields, which is challenging because input images that are wildly out of the distribution (e.g., images like vehicles, animals, humans, or a blurred image of an insect or insect class that is not yet trained on) can produce an incorrect insect classification. Out-of-distribution (OOD) detection algorithms provide an exciting avenue to overcome these challenge as it ensures that a model abstains from making incorrect classification prediction of non-insect and/or untrained insect class images. 

We generate and evaluate the performance of state-of-the-art OOD algorithms on insect detection classifiers. These algorithms represent a diversity of methods for addressing an OOD problem. Specifically, we focus on extrusive algorithms, i.e., algorithms that wrap around a well-trained classifier without the need for additional co-training. We compared three OOD detection algorithms: (i) Maximum Softmax Probability, which uses the softmax value as a confidence score, (ii) Mahalanobis distance-based algorithm, which uses a generative classification approach; and (iii) Energy-Based algorithm that maps the input data to a scalar value, called energy. We performed an extensive series of evaluations of these OOD algorithms across three performance axes: (a) \textit{Base model accuracy}: How does the accuracy of the classifier impact OOD performance? (b) How does the \textit{level of dissimilarity to the domain} impact OOD performance? and (c) \textit{Data imbalance}: How sensitive is OOD performance to the imbalance in per-class sample size?
Evaluating OOD algorithms across these performance axes provides practical guidelines to ensure the robust performance of well-trained models in the wild, which is a key consideration for agricultural applications. Our results indicate that OOD detection algorithms can significantly enhance user trust in insect pests classification by abstaining from performing classification under uncertain conditions.
\end{abstract}

\section*{Abbreviations}


\begin{center}
    
\begin{tabularx}{0.5\textwidth} { 
  | >{\raggedright\arraybackslash\hsize=2cm}X 
  | >{\centering\arraybackslash}X 
  |  }
 \hline
 OOD & Out-of-distribution  \\
 
 \hline
 ID & In-distribution  \\
 
 \hline
 MSP & Maximum Softmax Probability \\
 \hline
 MAH & Mahalanobis Distance \\
 \hline
 EBM  & Energy-Based Models  \\
\hline
\end{tabularx}
\end{center}

\clearpage

\section{Introduction}

Insect pests infestation can be observed at all stages of growth in crop plants, negatively affecting the quality and quantity of yields in agriculture \cite{skendvzic2021impact}. Accurate detection of insects is imperative for prompt, timely, and optimal decision-making \cite{dent2020insect}. For instance, early detection of pest infestations can allow farmers to take timely and appropriate action to prevent or minimize crop damage \cite{noar2021early,heeb2019climate}. More importantly, accurate detection allows farmers to identify the specific pest species that are causing damage \cite{kim2019advances}, enabling them to use targeted pest control methods instead of blanket applications of pesticides. This reduces the risk of harm to beneficial insects and other non-target organisms. Furthermore, accurate identification of insect pests can result in effective pest control measures which reduce the number of crop losses due to insect pests, increase the profitability of farmer operations, and reduce the amount of chemical runoff into water bodies. Finally, accurate and timely insect pest detection is also important for compliance with regulatory requirements related to pesticide use and environmental protection. 

Traditionally, insect-pest identification (and quantification)~\cite{naik2017machine, singh2016machine,xia2018insect} has been performed by human experts and scouts. In the past decade, machine learning approaches have increasingly been used for automating plant stress identification, classification, and quantification (Singh etal 2016, singh etal 2018). Researchers initially used classical machine learning algorithms for insect identification and classification~\cite{kasinathan2021insect,xia2018insect,chen2014flying}, however more recently, deep learning approaches have been suggested \cite{hoye2021deep, feuer2023zero,kar2021self}. In the classical approaches, a set of features such as color and texture are first extracted from images, and then a classifier learns a mapping from the feature space to their corresponding label. For example, support vector machines (SVM) have been commonly used for plant stress detection problems~\cite{ebrahimi2017vision, kasinathan2021insect}. However, such approaches require extensive domain knowledge \cite{singh2016machine,singh2018deep}, and tedious approaches for feature extraction. In the past decade or so, researchers have utilized deep-learning approaches to perform end-to-end classification, without the need for manual feature extraction. In classifying 13 soybean pests, Tetila et. al~\cite{tetila2020detection} compared the performance of 5 models (Inception-v3 \cite{szegedy2016rethinking}, ResNet50 \cite{he2016deep}, VGG16 \cite{simonyan2014very}, VGG19 \cite{simonyan2014very}, and Xception \cite{chollet2017xception}) across a dataset containing 5000 samples. Li et al.~\cite{li2020crop} leveraged the GoogLeNet architecture and achieved 98 percent accuracy in a 10-class insect classification task using a dataset of around 7000 images. Manual labeling of this moderately sized dataset required extensive effort by domain experts. The need for manual annotation of large datasets for training deep learning algorithms has become a major bottleneck to creating high accuracy classifiers. Recent efforts to utilize self-supervised algorithms seek to eliminate the need for large, annotated datasets. Using self-supervised approaches, one can eliminate the need for labeling by utilizing specific property of the data to design pretext task where model learns the underlying structure and relationships between elements in the input data while performing the pretext task.  In a recent study by Kar et al.~\cite{kar2021self} the authors utilized the  Bootstrap Your Own Latent, a self-supervised method to create a pest classifier algorithm. This insect classifier achieved an accuracy of 93 percent using approximately 15,000 images spanning 22 classes of agriculturally relevant insects found in the US Midwest. A review of field highlights a growing trends towards  utilizing largerdatasets such as the iNaturalist dataset (weblink) with $>$ 10 million images), along with weakly- and self-supervised approaches to develop high accurate classifiers for a wide range of agriculturally relevant insect-pests species. This progress opens up the possibility for the practical deployment in agricultural decision support systems. 

\begin{figure*}[t]
    \centering
    \includegraphics[width=0.8\textwidth, trim={0mm 0mm 0mm 0mm},clip]{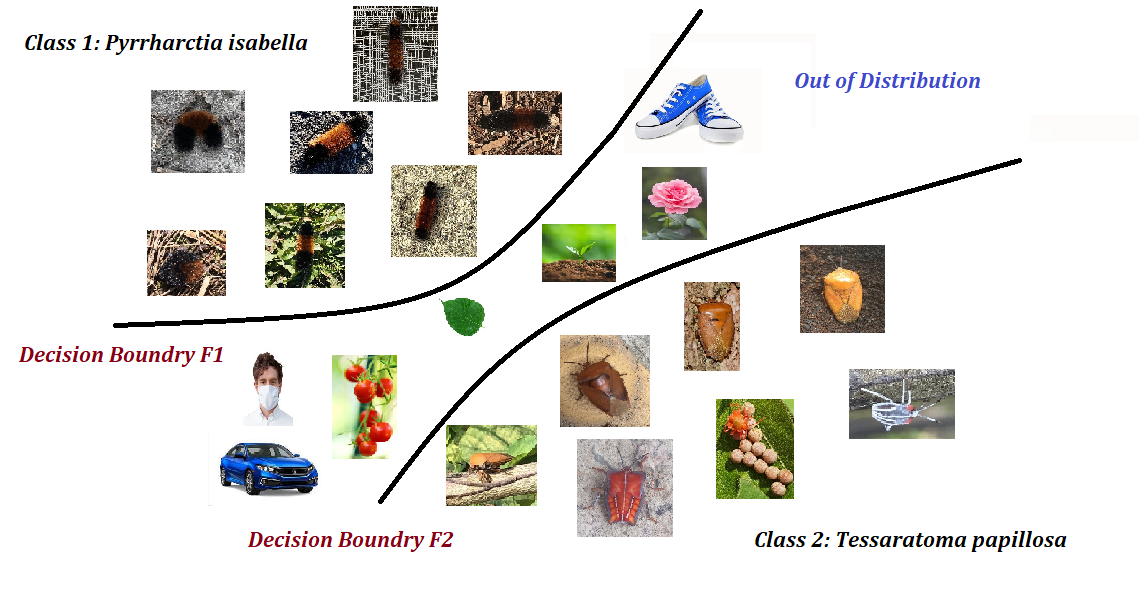}
    \caption{Out-of-distribution visualization in insect classification. The OOD classifier tries to define decision boundaries for each insect while ensuring no non-insect images fall into any known insect classes. There is a spectrum of OOD images. Here, we distinguish them as \textit{contextually near} OOD (for instance, leaf images) and \textit{contextually far} OOD (for instance, car image) based on their similarity to the in-distribution (ID) images.
        \label{fig:fig1}}
\end{figure*}

However, the trustworthiness of these classifiers is a potential bottleneck for their practical deployment in agricultural fields. For instance, while classifiers (or well-trained ML models in general) can achieve high classification accuracy, these models do not provide any measure of (un)certainty in making predictions or abstaining from classification in the case of uncertainty. This holds particularly important for autonomous agricultural applications, inaccurate predictions may lead to wastage of resources at the very least and in the worst-case scenario, substantial yield losses. In the context of ML models for insect classification, practical deployment requires distinguishing--- or at least abstaining from making predictions--- when confronted with confusing or unseen images. This can occur when the model is presented with images of insect species that do not fall under any of the categories on which it has been trained. Such misclassification could have disastrous consequences if the model mistakenly identifies an unseen insect class, for instance, an invasive species of insect as a harmless one. This may result in incorrect crop management decisions. Although the sizes of such confusing classes may vary, the loss of scale perspective in the image may contribute to these erroneous predictions. Thus, there is a pressing need to integrate advances in model accuracy with enhanced trustworthiness to ensure robust deployment in the field \footnote{Trustworthiness can be defined in several ways. See ~\cite{duncan2022veridicalflow, toreini2020relationship,gadiraju2020can}}. Increasing the trustworthiness of ML models is an active area of research with several promising approaches~\cite{bhatt2021uncertainty,le2022trustworthy,chatzimparmpas2020state}. One approach is to identify images that do not belong to the data distribution the model originally trained on. Such out-of-distribution (OOD) approaches can ensure that the model can abstain from making predictions when facing an image that does not belong to any of the classes it is trained to predict. This is particularly attractive in the context of insect pest detection as it allows human intervention in case of uncertain model predictions. Fig.~\ref{fig:fig1} illustrates how such OOD detection can be used for the insect classification task.

OOD detection has been successfully used in several applications --- like autonomous systems~\cite{meyer2020learning,cai2020real}, medical diagnosis \cite{cao2020benchmark,karimi2020improving,linmans2020efficient}, robotics ~\cite{farid2022task,yuhas2021embedded}, and social science~\cite{boyer2021out}--- to improve the reliability and safety of systems. In autonomous systems, OOD detection can help self-driving cars avoid any action or stop when they interface with an unknown object, which is crucial for the safety of passengers and pedestrians. For deep learning models, one of the earliest algorithms proposed for identifying out-of-distribution data is \textit{maximum-softmax-probability} (MSP)~\cite{hendrycks2016baseline}. This algorithm relies on the assumption that deep learning models are more confident in the classification of in-distribution data rather than out-of-distribution data. The algorithm uses the softmax value as a metric to measure the confidence of predictions. Due to its simplicity and good performance, this algorithm has been prevalent in addressing OOD detection. However, it has been shown in practice that MSP can produce a high false positive rate during OOD detection. Recently, Liu et. al.~\cite{liu2020energy} showed (using both theory and empirical evidence) that an \textit{energy-based} model can be a great substitute for MSP as they are aligned with the probability density of in-distribution data. The basic idea here is to design a function that maps the input (or input features) to a number, often called energy. This energy function is designed such that in-distribution samples have low energy while out-of-distribution samples exhibit higher energy. OOD boils down to comparing the energy of an unseen input against a threshold.

In contrast to these discriminative OOD algorithms, which try to find the best decision boundaries, there are also OOD algorithms based on generative models that focus on estimating the in-distribution density~\cite{choi2018generative,nalisnick2018deep,ren2019likelihood,serra2019input}. Among generative model-based OOD algorithms, OOD detection based on \textit{Mahalanobis distance} is the most popular, with significant recent work. Denouden et.al.~\cite{denouden2018improving} solves the OOD problem in the context of auto-encoder architecture, using the observation that an auto-encoder is ineffective in encoding and reconstructing OOD data in comparison to in-distribution ones. They distinguish OOD data from in-distribution by defining a threshold based on the Mahalanobis distance metric on reconstruction error. Lee et.al.~\cite{lee2018simple} extracted class conditional Gaussian distributions of deep learning features based on Gaussian discriminant analysis, leading to a Mahalanobis distance-based confidence score. Ren et.al.~\cite{ren2021simple} commented that the latter algorithm suffered from near-OOD detection and offered an adjustment to the previous algorithm. 

All the algorithms described above are \textit{extrusive}; i.e. they wrap around a trained ML classifier without the need for additional (re)training of the classifier. Another class of OOD algorithms are available that work by introducing part of the OOD data during the classifier training process. These \textit{intrusive} approaches require co-training of the OOD detection along with classifier. For instance, Hendrycks et.al.~\cite{hendrycks2018deep} modify the classifier's cross-entropy loss function and add an extra term to handle OOD data so that the softmax distribution for OOD data is uniform. Researchers also incorporated OOD detection into the classifier's architecture and designed a hierarchical outlier detection algorithm (HOD) to identify the outlier dermatological conditions~\cite{roy2022does}. Fort et.al.~\cite{fort2021exploring} and Ren et.al.~\cite{ren2019likelihood} also adjust the classifier by adding an extra class to classify the OOD data during the prediction.

We choose to focus on the extrusive algorithms from the agricultural application point of view, for the following reasons: (a) The increasing availability of well-trained models for various agricultural phenotyping tasks. These models are often trained on huge datasets using significant computational effort. Retraining them to incorporate OOD detection using intrusive algorithms may be computationally infeasible for many practitioners. (b) Additionally, extrusive algorithms allow efficient \textit{personalization} of the OOD detection for the specific application cases involved. 

In this paper, we explored the utility of OOD models for agricultural applications, using a specific application of insect pest classification. This work addresses a gap to enhance trustworthiness in the agricultural domain, particularly for insect pest classification. We perform OOD analysis on an insect classifier that is able to distinguish between 142 agriculturally relevant insect classes. This large class size makes OOD detection particularly relevant. Our key contributions are as follows: 
\begin{itemize}
\item We incorporate the concept of OOD detection for insect-pest classification, using a classifier that can distinguish between 142 agriculturally relevant insect pest species trained on a large insect dataset (2 million images). This differs from previous works, which conduct their analysis on benchmark datasets such as CIFAR10 \cite{krizhevsky2009learning}, CIFAR100 \cite{krizhevsky2009learning}, and SVHN \cite{netzer2011reading}, which have relatively smaller data sizes ( $\le$ 100,000 samples). 

\item We computationally explore the performance of three OOD methods (MSP, Mahalanobis distance, and energy-based model) for insect-pest classification. We report on OOD performance using different classifier architectures, different training dataset distributions, and testing on different OOD datasets. 
\end{itemize}

The paper is structured as follows. Section~\ref{Sec:2} provides an overview of the problem definition, including a discussion of the OOD detection methods, in-distribution and out-of-distribution datasets, and classifier architectures. Section~\ref{Sec:4} details the set of analyses that were conducted to evaluate OOD detection algorithms through different performance axes. Finally, in Section~\ref{Sec:5}, concluding remarks are provided, along with a discussion of future works.

\section{Materials and Methods}\label{Sec:2}

To understand the characteristics of OOD algorithms in the context of insect-pest classification, we selected three OOD algorithms - Maximum Softmax Probability, Mahalanobis distance, and energy-based model. Maximum Softmax Probability serves as a suitable baseline for comparison, while the Mahalanobis distance-based and energy-based algorithms each represent distinct families of OOD algorithms, discriminative and generative, respectively. We chose extrusive OOD algorithms as they can be applied to any arbitrary insect classifier as a wrapper. For the performance evaluation of these OOD detection algorithms, we perform a comprehensive set of analyses, as illustrated in Fig.~\ref{fig:fig2}. We first prepared four OOD datasets and the ID dataset with 142 insect classes. Next, we used various insect classifiers that used diverse architectures and finally applied the OOD algorithms on the combination of insect classifiers and datasets and measured their performance. We detail each of these steps next (OOD algorithm description $\rightarrow$ classifier architecture description $\rightarrow$ data description), starting with the description of the algorithms; see also Fig.~\ref{fig:fig3} for a schematic of these algorithms. All our experiments were performed on the local high-performance computing cluster, using one NVidia A100 80GB GPU card.



\subsection{Out-of-distribution Methods}  
Out-of-distribution (OOD) detection refers to the process of identifying whether a given input data point belongs to the same distribution as the training data or not. In other words, it aims to identify when a model encounters data that is significantly different from what it has been trained on. We detail the three OOD algorithms considered:

\begin{figure*}[h]
    \centering
    \includegraphics[width=.9\textwidth, trim={0mm 0mm 0mm 0mm},clip]{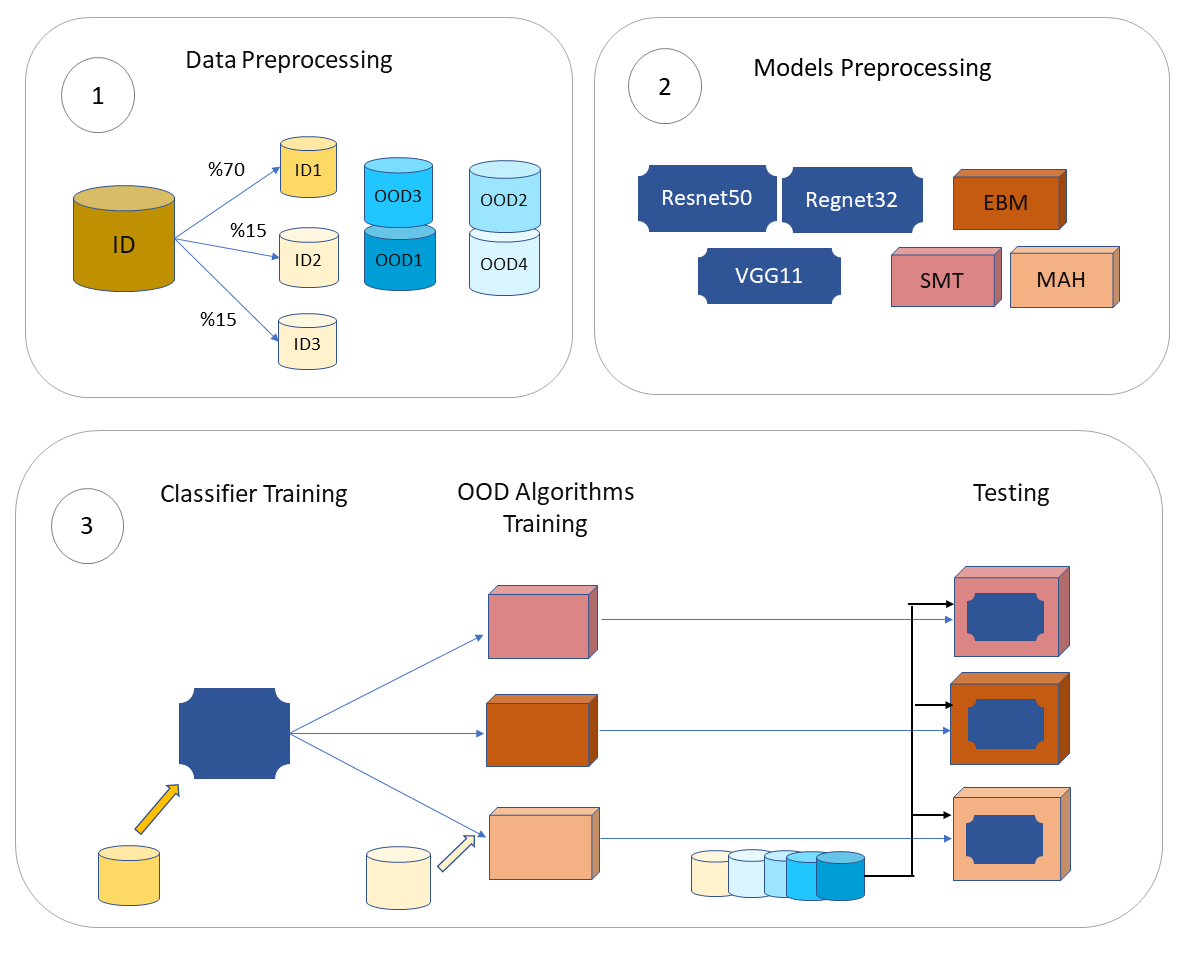}
    \caption{The analysis process: Box 1 represents the data. We split the ID-distribution data into three sub-datasets that are used for training the insect classifier, training the Mahalanobis distance OOD algorithm, and testing the results. We also utilize four different OOD datasets for testing our results. Box 2 shows the three insect classifier architectures, along with the three OOD algorithms used. Box 3 illustrates the workflow. The training data (ID1) is used to train the insect classifier. The OOD algorithms (and ID2 data, if needed) are next trained. Finally, the test data (ID3, along with OOD1, OOD2, OOD3, and OOD4) is used to test the OOD performance.\label{fig:fig2}}
\end{figure*}

\begin{itemize}
    \item{\textbf{Maximum Softmax Probability(MSP)} \cite{hendrycks2016baseline}} 
    is a simple yet effective method for OOD detection in deep learning models. It is based on the fact that softmax output probabilities (the last layer of the deep learning network before classification) for in-distribution data tend to be more confident than those for OOD data. Specifically, the MSP method computes the maximum value of the softmax probabilities for a given input data point and compares it to a threshold value. If the maximum value is below the threshold, the data point is considered OOD. The threshold is usually determined based on a validation set or using a predefined value.

    MSP has been shown to perform well in detecting OOD data in various applications, such as image classification, natural language processing, and speech recognition \cite{hendrycks2016baseline}. It is also computationally efficient and easy to implement as it only uses softmax value extracted from the last layer of the neural network, making it a popular choice for OOD detection in practice \cite{bevandic2018discriminative,chen2021atom,lee2018simple}. However, MSP may not perform well when the distribution of in-distribution data is significantly different from the training data or when there is an overlap between in-distribution and OOD data distributions. In such cases, more sophisticated methods, such as Mahalanobis distance or energy-based models, may be more effective.
    
    \item{\textbf{Mahalanobis distance-based algorithm} \cite{lee2018simple}} is another commonly used method for OOD detection in deep learning models. Mahalanobis distance measures the distance between a given input data point and the distribution of in-distribution data in the \textit{feature space}. It uses a distance metric that considers the correlation between features and the covariance matrix of the training data. 

    The feature space is simply the output of the penultimate layer of the classifier, denoted as $f(x)$. Given a classifier with labels, $\{1,...,C \}$, we first estimate the spread of the features for each class. This is estimated by assuming that these features belong to multivariate Gaussian distributions. Therefore, the probability of a feature vector, $f(x)$, given an input, $x$ with a class label $y =c$ is given by $P(f(x)|y=c) = \mathcal{N}(f(x)|\mu_c,\hat{\Sigma})$ where $\mu_c$ and $\hat{\Sigma}$ are respectively the mean and covariance.

    The Mahalanobis distance-based (MAH) confidence score, $M(x)$, is then the distance (of the feature, $f(x)$) of an unseen input $x$ to the \textit{closest} class-conditional Gaussian distribution:
    \begin{equation}
     M(x) = \max\limits_c -(f(x) - \hat{\mu_c})^T \hat{\Sigma}^{-1}(f(x)-\hat{\mu_c})     
    \end{equation}
    The Mahalanobis distance is the normalized distance, with the normalization performed using the covariance matrix of the training data. This helps account for correlations between features and results in a more accurate distance measure.
    A threshold for this distance is computed using a given OOD training sample $\{ (x_1,y_1),..., (x_N, y_N)\}$

    \item{\textbf{Energy-based models (EBM)} \cite{liu2020energy}} are a type of probabilistic model that assigns an energy score to the feature vector, $f(x)$, of each input, $x$. Using this energy score, an EBM can be used to distinguish between in-distribution (ID) and OOD samples. During inference, the energy score of an unseen input is compared to the energy scores of the training data. If the energy score of the input is significantly higher than the highest energy score of the training data, it is likely an OOD sample. 
    The energy model uses the softmax representation, $f(x)$ of a trained classifier, and the energy function is defined as  
        \begin{equation}
        E(x)= -T log \left( \sum_{i}^{C} e^{f_i(x)/T } \right) 
        \end{equation}
    The form of the energy function is motivated by other energy-based models like Boltzmann machines and variational autoencoders, where the energy function is related to the Gibbs distribution and the associated free energy. In the above formula, the parameter $T$ is called the temperature, which is set to $T=1$.


\end{itemize}

 \begin{figure*}[t]
        \centering
        \includegraphics[width=0.7\textwidth, trim={0mm 0mm 0mm 0mm},clip]{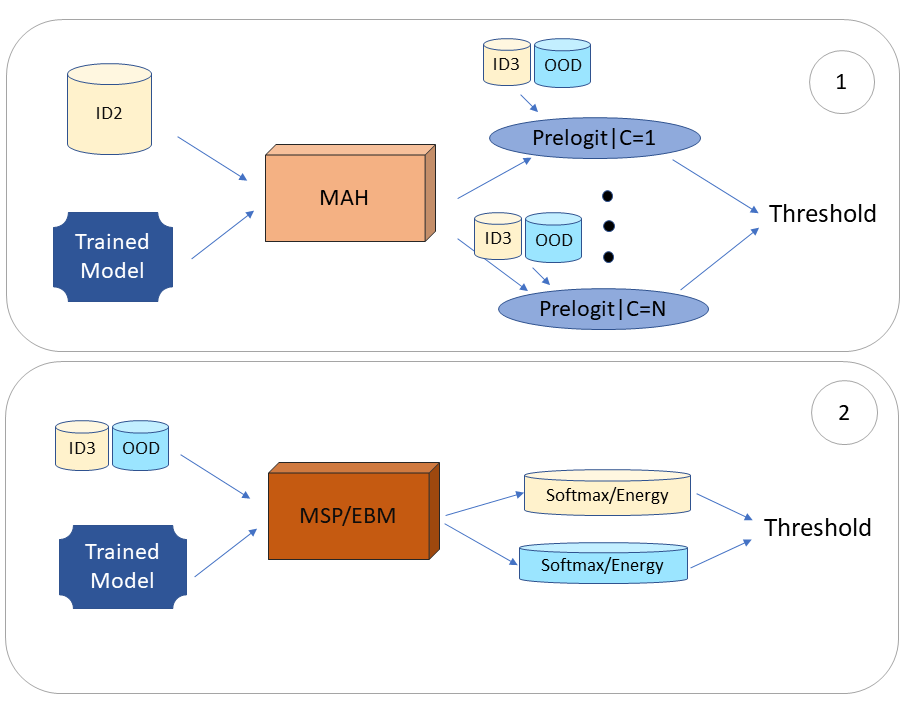}
        \caption{OOD detection process on MAH (top row), and MSP/EBM (bottom row). (1) In the Mahalanobis distance algorithm, the trained insect classifier and in-distribution data are given as input to the model, and the distribution of the feature vector (here, the prelogit) is conditioned on each class extracted. Then Mahalanobis distance of ID and OOD test data to the nearest conditional feature vector is calculated, and a threshold is obtained. For MSP and EBM instead of Mahalanobis distance, softmax, and energy are calculated, respectively.     
            \label{fig:fig3}}
\end{figure*}

\subsection{Insect Classifiers Methods}
We next describe the architectures used to train the insect classifier. The architecture of a classifier plays a critical role in determining the accuracy and efficiency of the model, and careful consideration and experimentation are often required to find the optimal design for a given task. We utilize three diverse types of architectures that have found significant success in practical image classification applications. These three network architecture choices, along with the three OOD algorithms described in the previous sub-section, provide a broad cross-section of approaches on which to evaluate OOD performance for our specific insect classification problem.

\noindent We consider the following architectures:
\begin{itemize}
    \item \textbf{ResNet50}:
 ResNet50, a variant of the convolutional neural network (CNN) model proposed by He et al.~\cite{he2016deep}, has gained widespread popularity for its exceptional performance in computer vision and image classification tasks. The model owes its success to incorporating skip connections within its residual blocks, effectively addressing the challenge of diminishing or exploding gradients. Our implementation employs ResNet50, featuring 50 deep convolution layers.

    \item \textbf{RegNetY32}:
    RegNet is an optimized design space developed by Radosavovic et al.~\cite{radosavovic2020designing} where they explore a diverse set of parameters of a network architecture like width, depth, groups (commonly called as AnyNet, an initial space of unconstrained models which uses models like ResNet50 ~\cite{he2016deep} as its base). By conducting numerous experiments with different parameter values for the design space, they successfully developed the optimized RegNetX. The combination of the well-known Squeeze-and-Excitation (SE) operation with ResNetX produced RegNetY, which exhibited outstanding performance.\cite{hu2018squeeze} We employed the RegNetY32 model in this study .
    \item \textbf{VGG11}:
     VGG is a CNN introduced by the Visual Geometry Group (VGG) at the University of Oxford~\cite{simonyan2014very}. This CNN architecture uses RELU activation function and pretrained weights (trained on ImageNet) to reduce computation load. We use VGG11 which is a variant of VGG. VGG11 is an architecture with 11 weighted layers. It includes 9 convolutional layers and three fully connected layers.
\end{itemize}

\subsection{Datasets}

\begin{figure*}[t]
    \centering
    \includegraphics[width=1\textwidth, trim={0mm 0mm 0mm 0mm},clip]{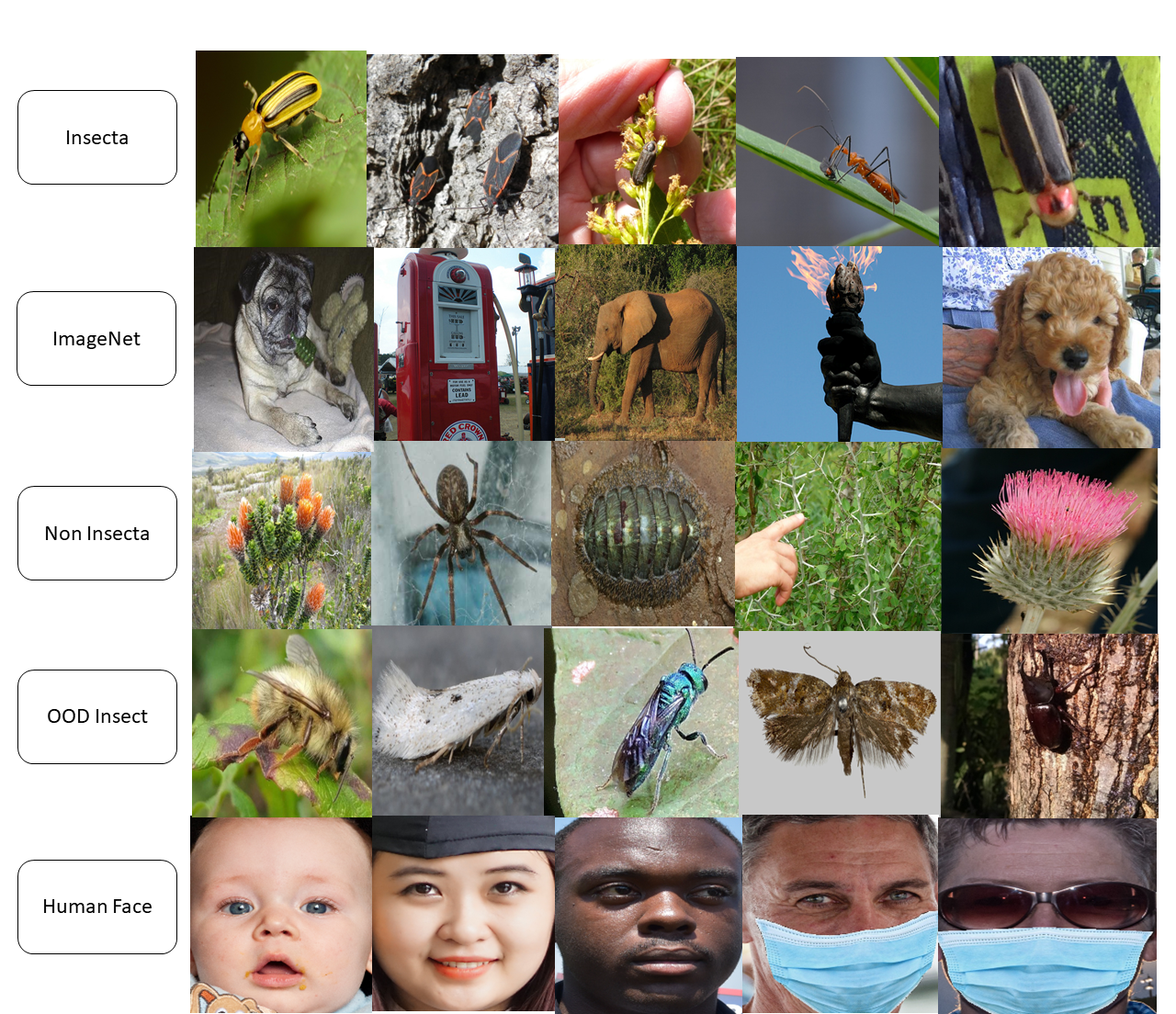}
    \caption{Dataset visualization.The Insecta dataset was used as In-Distribution and {ImageNet, Non-Insecta, OOD Insect, Human Face} were used as OOD datasets. 
        \label{fig:fig4}}
\end{figure*}

We curated two series of datasets, one for in-distribution (ID) and one for OOD. We used ID data for three main purposes: (i) training the insect classifier, (ii)  training the generative OOD algorithm (Mahalanobis distance), and (iii) evaluating the performance of OOD algorithms with respect to distinguishing the ID data from OOD data. For these objectives, we curated an insect dataset consisting of 142 agriculturally relevant species with a large economic impact in North American agriculture. This dataset is a subset of the publicly available iNaturalist dataset~\cite{Van_Horn_2021_CVPR} and consists of 2 million insect pictures. We split the dataset into train and validation sets with a ratio of 7 to 3, and use the training subset to train the classifier. We then divide the validation folder into two equally sized smaller datasets. One was used for training the OOD model (only for the generative OOD model), and the other was used for evaluating the performance of OOD algorithms.   

For OOD data, we utilized four datasets with different degrees of similarity to the in-distribution insect data. An illustration for the datasets are provided in Fig~\ref{fig:fig4}. We briefly describe each of these datasets.
\begin{itemize}
 
\item ImageNet~\cite{ILSVRC15} (far OOD):
We start with the ImageNet 2012 classification dataset consisting of 1.2 million images across 1000 object categories. We then took ten random samples from each class and excluded all insect-related objects (e.g. bee, cricket, ant) from it, which resulted in 9730 samples.

\item Human Face~\cite{wang2020masked} (far OOD):
We use the dataset from the face mask recognition dataset of the Kaggle competition\footnote{\url{https://www.kaggle.com/datasets/ashwingupta3012/human-faces}}, which includes pictures of human faces with and without masks. The dataset consists of 3059 images. 

\item Non-Insecta ~\cite{Van_Horn_2021_CVPR} (near OOD): this dataset is a subset of iNaturalist \footnote{\url{https://github.com/visipedia/inat_comp/tree/master/2021}},
 where we exclude all the Insecta images from it. This dataset consists of 74740 images.  
 
\item OODInsect (near OOD):
This dataset includes all insect pictures that do not belong to any of the 142 classes of ID data. This dataset is also collected from the iNatrualist dataset. This dataset includes 56487 pictures of OOD insects.
\end{itemize}

Taken together, these four datasets represent a diverse set of out-of-distribution images, when compared to the dataset on which the insect classifier is trained. 

\subsection{Metrices for OOD performance evaluation}
We employed AUROC and FPR95 to evaluate the performance of OOD detection algorithms. We briefly discuss each metric below:

\textbf{AUROC:} AUROC stands for the area under the receiver operating characteristics (ROC) curve, which is a performance measure for classification tasks, in this case distinguishing between ID or OOD. The ROC curve plots the true positive rate (TPR) versus the false positive rate (FPR) at various classification thresholds. The AUROC is the area under this curve, which ranges from 0 to 1. A higher AUROC indicates better OOD detection performance, as it means that the model has a higher TPR for a given FPR. AUROC is a useful metric for OOD detection because it captures the overall performance of a model at different operating points. By considering the full range of FPR and TPR values, AUROC provides a comprehensive evaluation of a model's ability to distinguish between ID (in-distribution) and OOD samples. However, it is important to note that AUROC does not provide information about the specific operating point at which the model achieves its highest performance. For this reason, additional metrics such as FPR95 are used to further evaluate the OOD detection performance.

\textbf{FPR95:} FPR95, or false positive rate at 95\% true positive rate, is a metric used to evaluate OOD (Out-of-Distribution) detection algorithms. FPR95 is the probability that a negative example (OOD) is wrongly classified as positive (ID) when the true positive rate (TPR) is equal to 95\%. A lower FPR95 indicates better OOD detection performance, as it means that the model is less likely to misclassify OOD samples as in-distribution (ID) samples. FPR95 is an important metric because it captures the performance of a model at a high TPR, which is where OOD detection is most critical.



\section{Results and Discussion}\label{Sec:4}

We performed a large number of computational experiments along three performance axes:
(i) classifier architectures and classifier accuracies, detailed in Section.~\ref{Result_Axis1} (ii) different sample sizes for training data, detailed in Section.~\ref{Result_Axis2}, and (iii) the degree of similarity of OOD dataset features to ID data, detailed in Section.~\ref{Result_Axis3}.

These experiments indicate that EBM outperforms the other two (MAH and MSP) OOD detection algorithms.
Additionally, our results indicate that the RegNetY32 architecture exhibits superior performance in insect detection compared to the other two model architectures (ResNet50, VGG11). 
Taken together, a combination of EBM with a RegNetY32 classifier produced an excellent insect classifier.

\begin{figure*}[ht!]
    \centering
    \begin{subfigure}[t]{0.49\textwidth}
        \centering
        \includegraphics[height=2.3in]{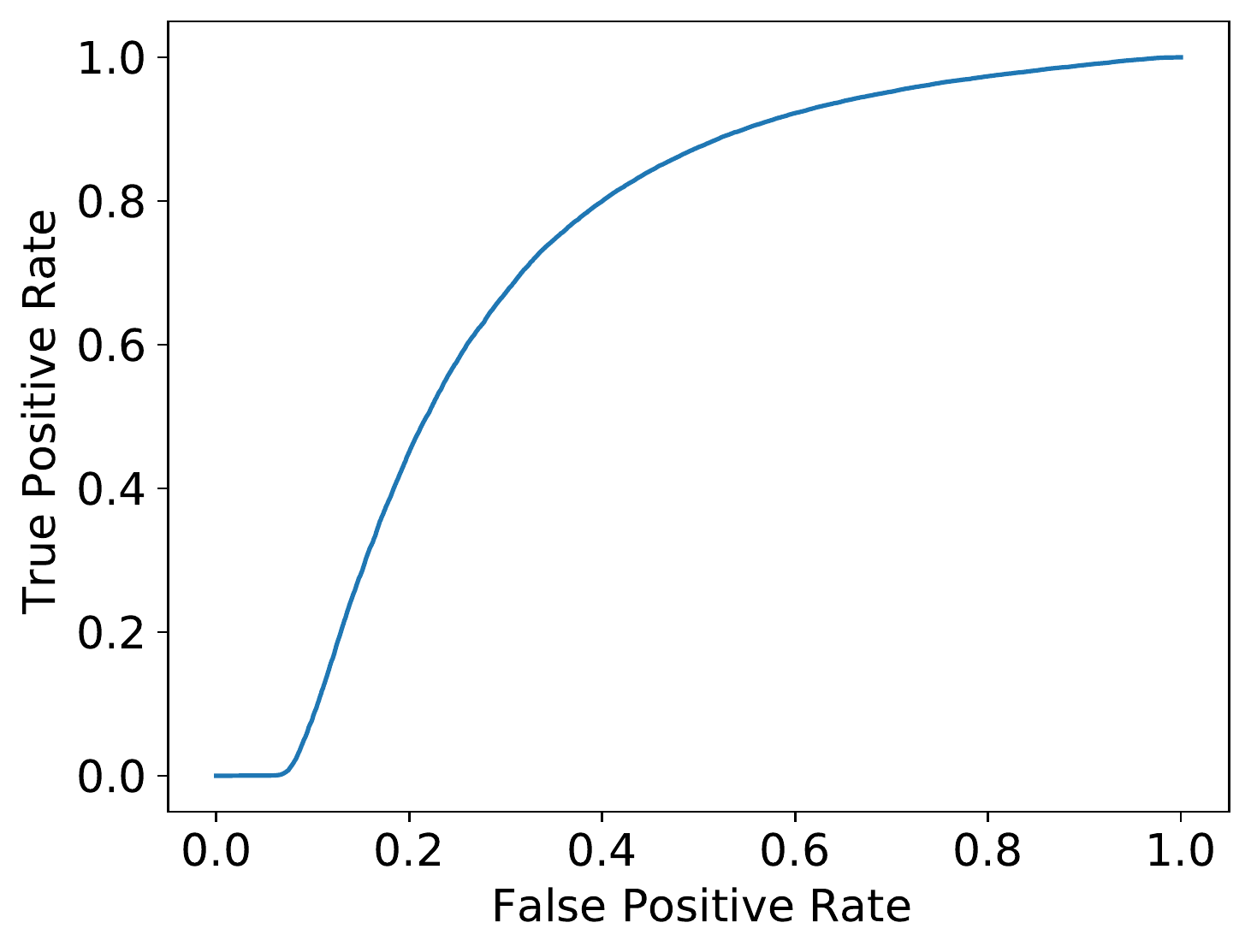}
        \caption{}
    \end{subfigure}%
    ~ 
    \begin{subfigure}[t]{0.49\textwidth}
        \centering
        \includegraphics[height=2.3in]{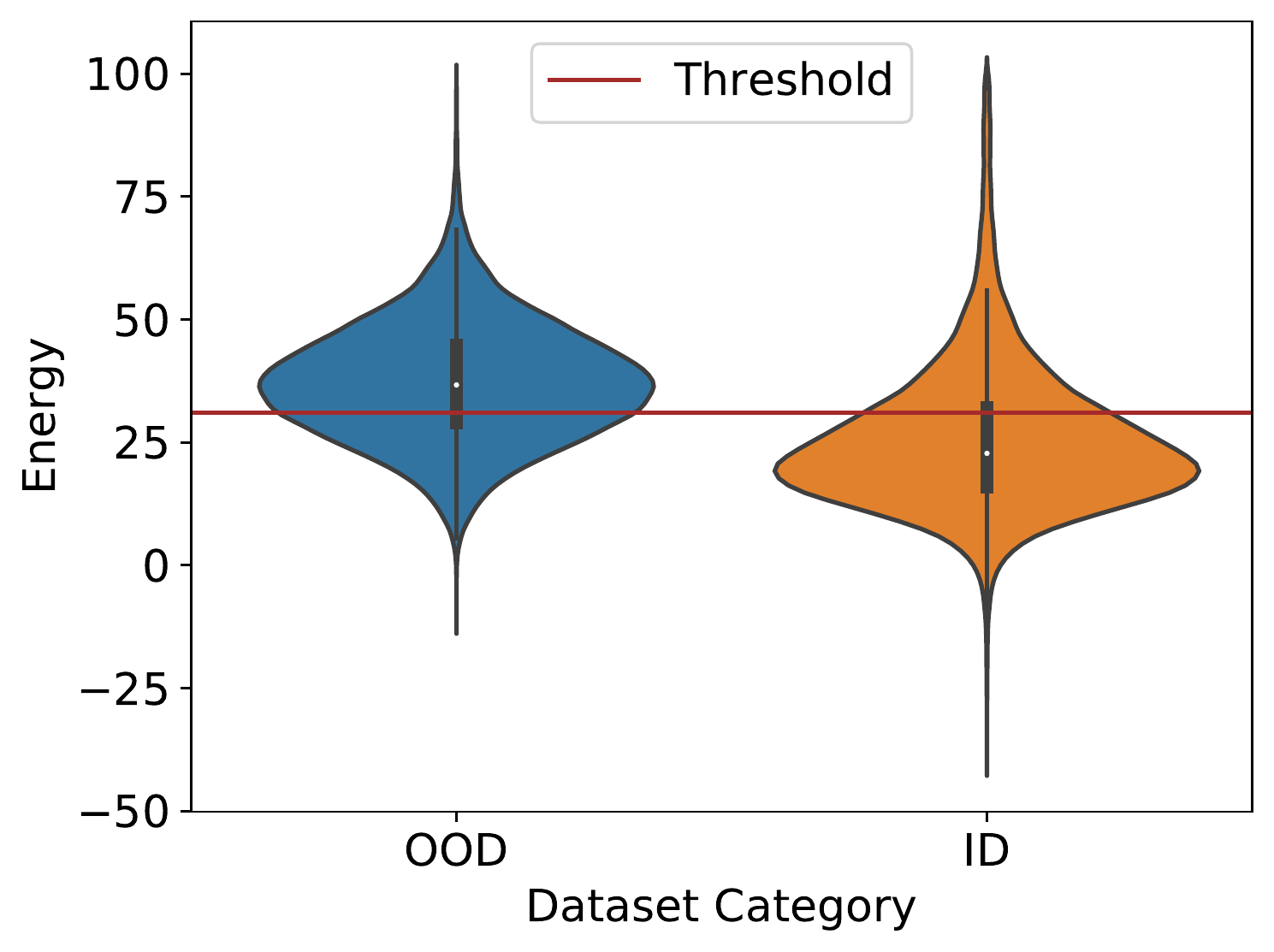}
        \caption{}
    \end{subfigure}
    \caption{The performance evaluation of EBM on RegNetY32. (a) AUROC curve of binary classification between ID and OOD for the range of energy thresholds in the proposed EBM OOD detector. (b) boxplot diagram of OOD and ID distribution.}
    \label{fig:fig5}%
\end{figure*}

We plot the AUROC curve for this classifier in  Fig.~\ref{fig:fig5}(a), which exhibits an aggregated OOD classifier performance over all possible energy level thresholds. The result of our analysis indicated an AUROC of 0.72. We then tried to find the threshold that gave us the most promising result. Based on our analysis, we found that the threshold of 31.0 gave the OOD classifier simultaneously the lowest FPR=0.31 and highest TPR=0.69, respectively ($FPR=0$ and $TPR=1$). Setting the energy level to this threshold resulted in an accuracy of 68.97 for our OOD model. 


In order to interpret the threshold and AUROC results, the energy distribution of ID and OOD data were further investigated. Fig.~\ref{fig:fig5}(b) captures the boxplot illustration of OOD and ID data distributions. They exhibit the following inter-quantiles: In-distribution inter-quantile of {\{min: -11.23, IQ1: 29.00, median: 36.68, IQ3: 44.71, max: 676.69\}}, and 
Out-of-distribution inter-quantile of \{min: -43.73,  IQ1: 16.33, median: 23.69, IQ3: 34.30, max: 2445.05\}. 

\noindent Majority of ID data were retained while an acceptable portion of OOD was excluded when using an energy threshold level of 31.0.  The wide scattering in the energy level of OOD data indicated the diversity of OOD images. Difficulty in OOD classification was also evident, as there was a partial overlap of low energy values on both ID and OOD.






\subsection{Impact of classifier accuracy (and architecture) on OOD detection}\label{Result_Axis1}

We explored the interplay between the accuracy of the insect classifier and the performance of the OOD algorithm. This question is motivated by a recent empirical observation~\cite{vaze2021open} that suggests that improving the accuracy of the classifier is correlated with its ability to identify when an input does not belong to any of its classes. This observation is particularly appealing because one could invest effort in improving the accuracy of the classifier (via improved data collection, augmentation, hyperparameter tuning, and label smoothing) and get the added benefit of improved OOD detection.

\begin{figure*}[t]
    \centering
    \includegraphics[width=1\linewidth]{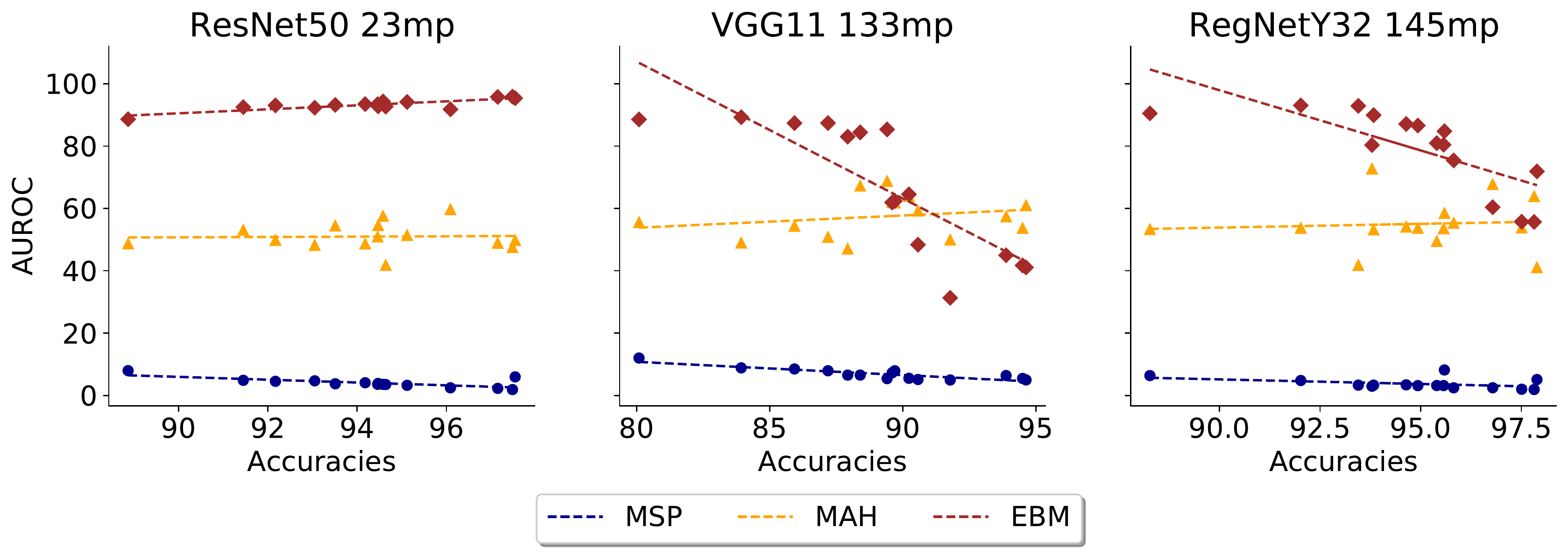}
    \caption{The trend of AUROC for the three OOD algorithms (MSP, MAH, EBM) over the increment in insect classifier's accuracy for the three architectures (ResNet50, RegNetY32, VGG11). Each diagram specifies the performance of one architecture, and diagrams are sorted from left to right based on the increase in the number of weight parameters ($mp$ stands for million parameters). }%
    \label{fig:fig7}%
\end{figure*}

To explore this, we created a sequence of 15 trained insect classifiers with increasing accuracy for each of the three types of network architectures (ResNet50, RegNetY32, and VGG11). These models are trained using cross-entropy loss function and AdamW optimizer with a learning rate of $1e-3$ and a batch size of $256$. Their performance -- on the validation dataset -- ranged from an accuracy of 80\% to 97.5\% across these 45 models, providing a statistically robust number of trained models. Next, we ran all three OOD algorithms (MSP, MAH, and EBM) on these 45 trained insect classifiers. For evaluating the OOD performance, we consider the non-Insecta dataset as the OOD dataset. We ensured both ID and OOD test datasets have the same size by random sampling of 74740 (the size of the Non-Insecta dataset) from ID data. The result of our analysis is shown in Fig.~\ref{fig:fig7}.

We observe several trends. First, the performance of the EBM algorithm is (almost always) better than the other two OOD algorithms across all three architectures. However, notice that improving the accuracy of the classifier \textit{does not} improve the EBM-based OOD detection performance, especially for architectures with a very large number of trainable parameters; VGG11 has 133 million trainable parameters, and RegNetY32 has 145 million trainable parameters. For architecture with a moderate number of parameters (i.e. Resnet50 with 23 million trainable parameters), the observation of \cite{vaze2021open} holds. On the other hand, the other two OOD algorithms (MSP and MAH) show slight to no improvement in OOD performance as classifier accuracy is increased. 


\subsection{Impact of \textit{degree of out-of-domainness} on OOD detection}\label{Result_Axis2}

In this sub-section, we explore how OOD performance is affected by the type of OOD datasets considered. This is particularly important to understand for agricultural applications, where the adoption of these tools may be compromised if the classifier makes obvious classification mistakes, for instance, classifying a human face as an insect.  

We considered four datasets -- Human face, ImageNet, Non-Insecta, and OODInsect -- that exhibit increasing contextual variations from the insect dataset used to train the classifiers. The non-Insecta and OOD Insect datasets share many similar features to the in-domain dataset, such as leaves or trees in the image background; on the other hand, ImageNet and HumanFace share less similar features to the ID insect dataset. These datasets are visualized in Fig.~\ref{fig:fig4}. Furthermore, to avoid potential bias from different sizes of datasets, we took a random sampling of 3059 (our smallest OOD dataset size) from each of these datasets. We then measured the performance of the three OOD algorithms on our best-performing classifier. 

\begin{figure*}[t]
    \centering
    \includegraphics[width=1\linewidth]{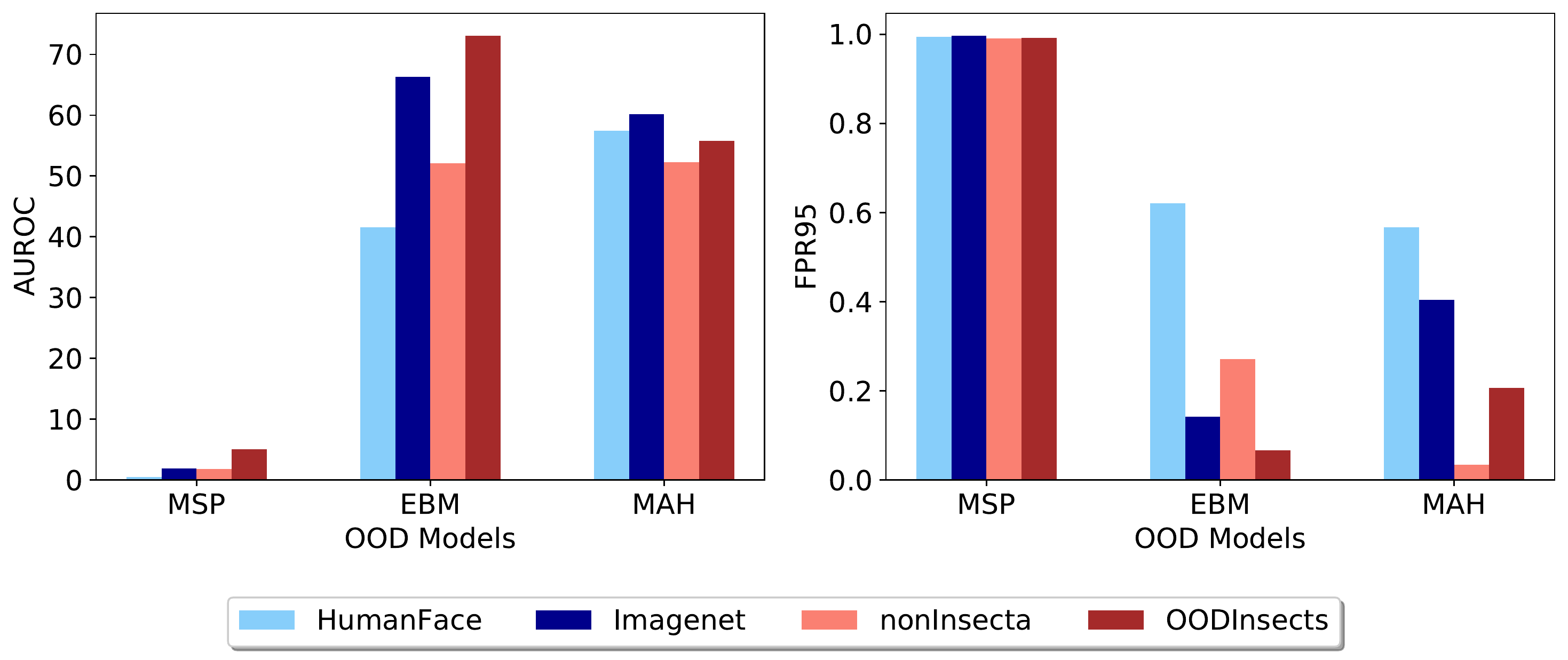}
    \caption{OOD detection algorithms performance across OOD datasets with different degrees of similarity to ID. Two metrics of AUROC and FPR95 are chosen to show two performance aspects. (Right) True Positive rate at False Positive rate 95\%, the lower the FPR95 the better performance of OOD detector.}
    \label{fig:fig8}%
\end{figure*}

Fig.~\ref{fig:fig8} lays out the results of this analysis. We observe that EBM has the best performance on all four OOD datasets -- exhibiting consistently good AUROC, as well as low FPR95.  MSP exhibited high FPR95, independent of the type of OOD data. This has been well documented ~\cite{liu2020energy}.  
EBM and MAH performed better than MSP on all OOD datasets. In particular, better performance was noted in the ImageNet dataset and OOD Insects. This could be because the classifier is pre-trained on Instagram images, which share similar features to the ImageNet dataset, and is trained on insect data that is similar to the OOD insect dataset. This makes the OOD detector more robust on these datasets, and suggests that the similarity of features between the images the classifier is trained (or pre-trained on) and the OOD samples can help improve the OOD detection. This observation raises the hypothesis that using models pre-trained on diverse/large datasets may produce especially robust OOD detection.  

\subsection{Effect of data imbalance on OOD detection}\label{Result_Axis3}

Here, we explore how data imbalance in the training data impacts OOD performance. We consider three subsets of ID data to train the OOD algorithms. 
We call these three datasets balanced, unbalanced, and unbalanced-uniform. For consistency, we ensured that each dataset had the same number of images (58362 images).
In the balanced subset, all classes have the same number of images (411 images per class). The unbalanced subset, on the other hand, exhibits the same data distribution as the original 2M image dataset used to train the classifier. Finally, in unbalanced-uniform, we considered a data distribution that is completely random, i.e. the size of each class is an independent uniform random variable. This allows us to explore the impact (or lack thereof) of specific data distributions exhibited by the unbalanced vs the unbalanced-uniform datasets. We compared these three ID subsets with the Non-Insecta OOD dataset on OOD detection performance on our best classifier.

\begin{figure*}[t]
    \centering
    \includegraphics[width=0.7\linewidth]{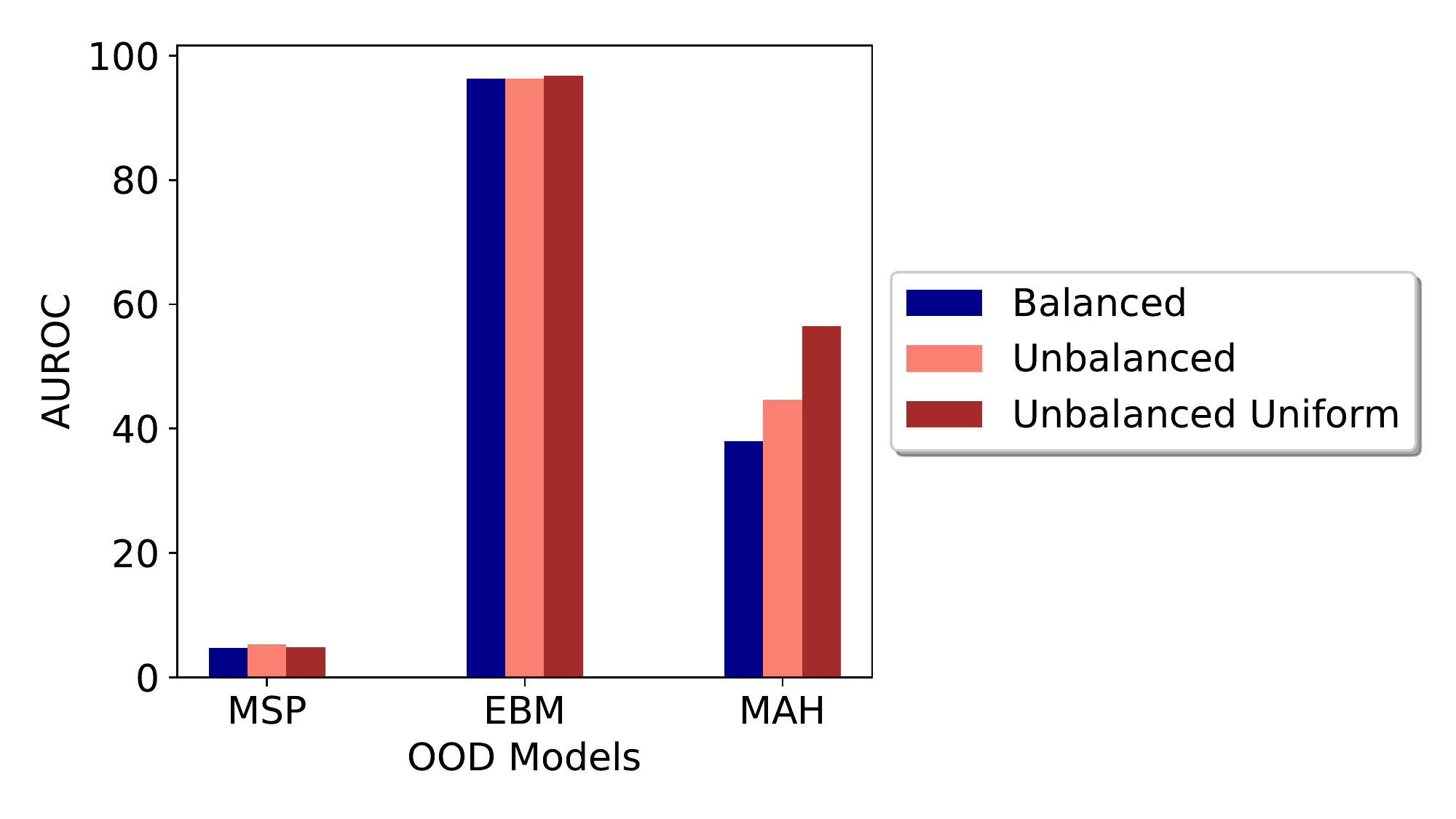}
    \caption{OOD performance across different levels of equilibrium of ID dataset. Only the MAH algorithm shows a significant difference in performance between these three subsets.}%
    \label{fig:fig9}%
\end{figure*}

The result of this analysis is shown in Fig.~\ref{fig:fig9}. EBM -- a discriminative OOD algorithm -- is robust to data imbalance, and performs very well with high AUROC. On the other hand, the generative OOD algorithm, MAH, shows sensitivity to data imbalance.
Over all, we observe that using EBM is particularly promising for agricultural applications, where there is often the possibility of data imbalance.

This analysis is especially important in agricultural applications, particularly for the insect classification problem, due to the possibility of significant data imbalance. For instance, charismatic species like the monarch butterfly are typically heavily imaged, while rare (or uninteresting) species can have very few samples in a dataset. One of the largest insect datasets, the iNaturalist dataset, consists of 13 million insect images. This dataset exhibits a large data imbalance between classes, with some classes containing fewer than 100 images and others containing more than 100,000 images. Such data imbalances are rather common in agricultural applications, and a clear understanding of the impact of such data imbalance on OOD performance is useful for practical deployment.   

\section{Conclusions}\label{Sec:5}
Automated insect pest identification is an economically critical agricultural task. It is important that well-trained models, when deployed in the wild, abstain from making predictions when encountering data that is out of their training distribution. We explored and quantified the performance of several OOD approaches applied to insect pests classification. Our insect pest classifier was trained on over 2 million images to accurately classify 142 agriculturally relevant insect species. For this particular application, the energy-based model (EBM) algorithm for OOD detection was the best-performing model. This work is the first step towards robustifying the performance of classifiers deployed for agricultural applications. 

We performed an extensive set of computational experiments to evaluate the performance of OOD algorithms, which provided several useful observations to the agricultural community. First, we find that for classifiers with fewer model parameters, there is a correlation between classifier performance and OOD detection accuracy. However, this correlation disappears for more complex models. Second, it was observed that the EBM OOD detector has superior performance when OOD datasets are more similar to the training and pretraining datasets. Third, empirical observations indicate that both discriminatively selected OOD detection algorithms, EBMs and MSP, exhibit insensitivity to data imbalance, in contrast to their generative counterpart algorithm, MAH, method.

The OOD approach described in this paper allows the classifier to abstain from making predictions when the input data does not belong to the training distribution. This provides a systematic approach for autonomous agricultural applications to request human intervention, thus improving the reliability and trustworthiness of these tools. We anticipate this approach to be applicable to a wide range of agriculturally relevant classification tasks, such as scouting and identification of biotic (disease) and abiotic stresses (nutrient deficiency), as well as providing assistance in agricultural decision-making processes.


\section*{Author Contributions}
    M.S., A.S., and B.G. conceived the project. Insect classifier training and data pre-processing were conducted by M.S. and S.C. OOD analysis and training was conducted by M.S. M.S., S.C, A.B., Z.J., and B.G wrote the manuscript draft. All authors reviewed and edited the manuscript. 

\section*{Acknowledgements}
This work was supported by the AI Institute for Resilient Agriculture (USDA-NIFA \#2021-67021-35329), COALESCE: COntext Aware LEarning for Sustainable CybEr-Agricultural Systems (NSF CPS Frontier \#1954556), FACT: A Scalable Cyber Ecosystem for Acquisition, Curation, and Analysis of Multispectral UAV Image Data (USDA-NIFA \#2019-67021-29938), Smart Integrated Farm Network for Rural Agricultural Communities (SIRAC) (NSF S\&CC \#1952045), and USDA CRIS Project IOW04714. Support was also provided by the Plant Sciences Institute.

\section*{Competing Interests}
    The authors declare that there is no conflict of interest regarding the publication of this article.

\printbibliography
\end{document}